%% file: main.tex
\documentclass[conference]{IEEEtran}
\usepackage{times}

\usepackage{multicol}
\usepackage{multirow}
\usepackage{mathtools}
\usepackage{amsthm}
\usepackage{graphicx}
\usepackage{amsmath}
\usepackage{amssymb}
\usepackage{booktabs}
\usepackage{array}
\usepackage{wrapfig}
\usepackage{multirow}
\usepackage[dvipsnames]{xcolor}
\usepackage{pifont}

\usepackage{color}
\usepackage[pagebackref=true,breaklinks=true,colorlinks,bookmarks=false,urlcolor=blue, citecolor=blue]{hyperref}

\usepackage[noadjust]{cite}

\hypersetup{
linkcolor=BrickRed
,citecolor=Green
,filecolor=Mulberry
,urlcolor=NavyBlue
,menucolor=BrickRed
,runcolor=Mulberry
,linkbordercolor=BrickRed
,citebordercolor=Green
,filebordercolor=Mulberry
,urlbordercolor=NavyBlue
,menubordercolor=BrickRed
,runbordercolor=Mulberry
}

\DeclareMathAlphabet{\pazocal}{OMS}{zplm}{m}{n}

\DeclareMathAlphabet\mathbfcal{OMS}{cmsy}{b}{n}
\newcommand{\acronym}{DeepExplorer}

\IEEEoverridecommandlockouts %<-- need this to show the \thanks in the \authorblockN below
\begin{document}

\title{Metric-Free Exploration for Topological Mapping \\by Task and Motion Imitation in Feature Space}
\author{\authorblockN{Yuhang He$^{1,}$\authorrefmark{1}\thanks{\authorrefmark{1} Equal contributions.} 
Irving Fang$^{2,}$\authorrefmark{1} 
Yiming Li$^{2}$ 
Rushi Bhavesh Shah$^{2}$ 
Chen Feng$^{2}$\textsuperscript{, \ding{41}}\thanks{\ding{41} Corresponding author. The work is supported by NSF grant 2238968.}
}
\authorblockA{$^{1}$ Department of Computer Science, University of Oxford, Oxford, United Kingdom~~\url{yuhang.he@cs.ox.ac.uk}}
\authorblockA{$^{2}$ Tandon School of Engineering, New York University, New York, United States~~\url{{zf540;yimingli;rs7236;cfeng}@nyu.edu}}
% \authorblockA{\url{https://ai4ce.github.io/DeepExplorer/}}
% \authorblockA{\authorrefmark{1}Department of Computer Science\\
% University of Oxford,
% Oxford, United Kingdom\\ Email: yuhang.he@cs.ox.ac.uk}
% \authorblockA{\authorrefmark{2} Tandon School of Engineering\\ New York University\\
% Email: zf540@nyu.edu; yimingli@nyu.edu; rs7236@nyu.edu; cfeng@@nyu.edu}
}

\maketitle
\input{abstract}
\IEEEpeerreviewmaketitle
\input{intro}
\input{RelatedWork}
\input{mainsec1_activeexp}
\input{mainsec2_topomapping}
\input{experiment}
\input{conclusion}
 
\bibliographystyle{unsrt}
\bibliography{references}
\clearpage
\input{suppmaterial}

\end{document}

%% file: abstract.tex
\begin{abstract}
We propose~\textit{\acronym}, a simple and lightweight metric-free exploration method for topological mapping of unknown environments. It performs task and motion planning~(TAMP) entirely in image feature space. The task planner is a recurrent network using the latest image observation sequence to hallucinate a feature as the next-best exploration goal. The motion planner then utilizes the current and the hallucinated features to generate an action taking the agent towards that goal. The two planners are jointly trained via deeply-supervised imitation learning from expert demonstrations. During exploration, we iteratively call the two planners to predict the next action, and the topological map is built by constantly appending the latest image observation and action to the map and using visual place recognition~(VPR) for loop closing. 
The resulting topological map efficiently represents an environment's connectivity and traversability, so it can be used for tasks such as visual navigation. We show \textit{\acronym}'s exploration efficiency and strong sim2sim generalization capability on large-scale simulation datasets like Gibson and MP3D. Its effectiveness is further validated via the image-goal navigation performance on the resulting topological map. We further show its strong zero-shot sim2real generalization capability in real-world experiments. The source code is available at \url{https://ai4ce.github.io/DeepExplorer/}.
\end{abstract}

%% file: intro.tex
\section{Introduction}
Mobile agents often create maps to represent their surrounding environments\,\cite{slam_summary}. Typically, such a map is either topological or metrical\,(including hybrid ones). We consider a topological map to be metric-free, which means it does not explicitly store global/relative position/orientation information with measurable geometrical accuracy~\cite{kuipers1991robot,topomap_sonarvision}. Instead, it is a graph that stores local sensor observations, such as RGB images, as graph nodes and the spatial neighborhood structure (and often navigation actions) as graph edges that connects observations taken from nearby locations. While metric maps are often reconstructed by optimizing geometric constraints between landmarks and sensor poses from classic simultaneous localization and mapping (SLAM), topological maps have recently attracted attention in visual navigation tasks due to the simplicity, flexibility, scalability, and interpretability~\cite{savinov2018semiparametric,neural_topomap,nav_maze,VisualGraphMem_ICCV21,learn2explore_iclr20,TSGM}.

\begin{figure}[t]
    \centering
    \includegraphics[width=0.98\linewidth]{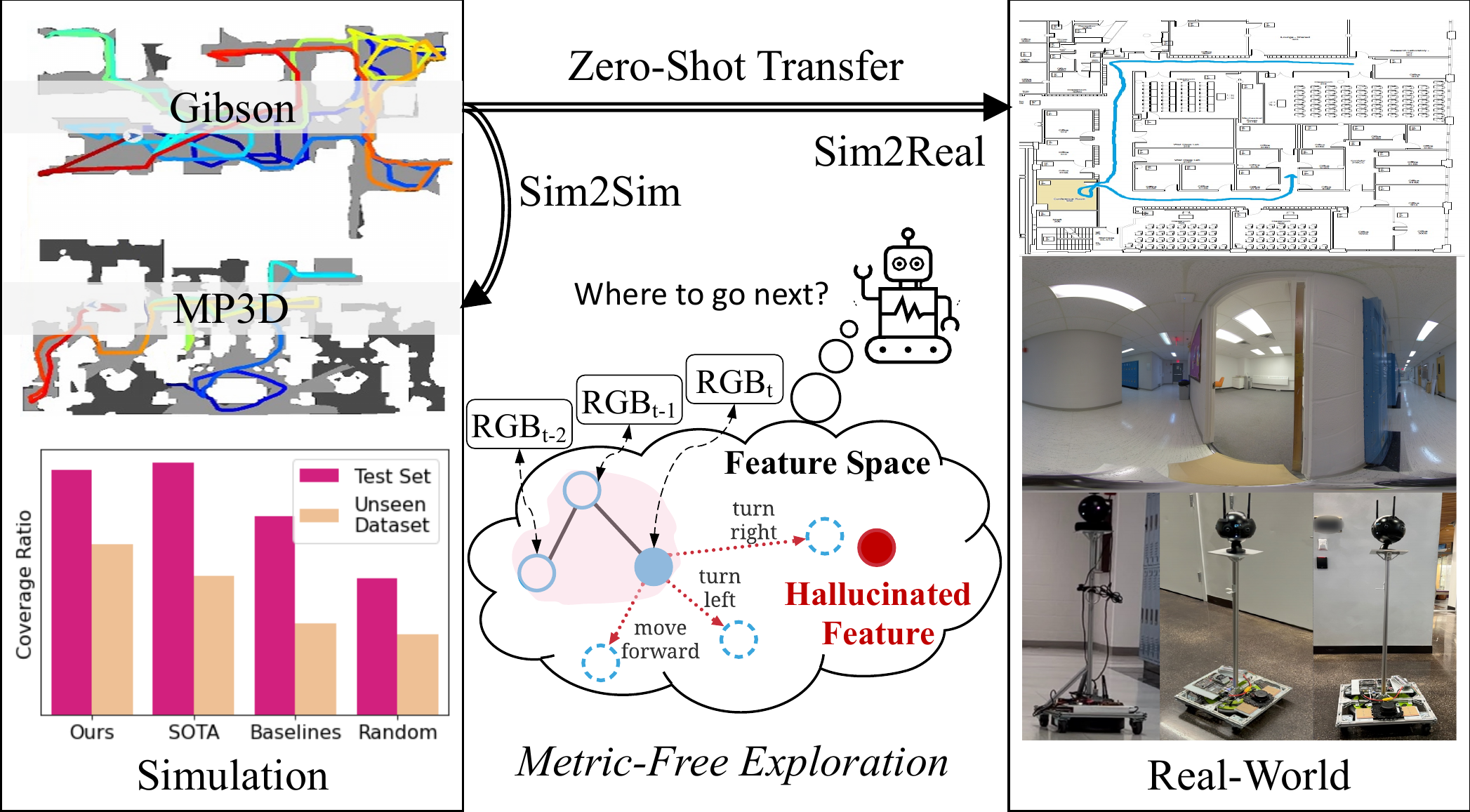}
    \vspace{-2mm}
    \caption{\textit{\acronym{}} illustration: it is metric-free and plans in the feature space. It jointly hallucinates the next step feature to visit and predicts the appropriate action taking the agent to the hallucinated feature. We show that \textit{\acronym{}} is efficient in simulation environment and possesses strong zero-shot sim2real capability in exploring real-world environment.}
    \label{fig:teasingfig}
    \vspace{-4mm}
\end{figure}

There are two robot exploration methods to collect data to construct a topological map in a new environment. 
The first and also the simplest one is to let the agent explore the new environment through metric-free \emph{random walk}, after which the topological map could be built by projecting the recorded observations into a feature space and adding graph edges from temporal connections and loop closures~\cite{savinov2018semiparametric}. However random walking is very inefficient especially in large or complex rooms, leading to repetitive visits to the nearby local areas. The other way is to design a navigation policy that controls the agent to more effectively explore the area while creating the map. It is known as \emph{active SLAM} and often involves some metric information (e.g., distance and orientation) from either additional input modalities~\cite{leung2006active,learn2explore_iclr20} or intermediate estimations~\cite{neural_topomap}. 
Could we combine the merits of the two ways by finding \emph{an exploration policy that (1) is metric-free thus simple and lightweight in hardware and model complexity, and (2) exhibits strong generalization ability to explore unknown environment} for topological map construction?

To achieve this objective, we propose \textit{\acronym{}} (see Fig.~\ref{fig:teasingfig}), a new framework to achieve metric-free efficient exploration by imitating easy-to-access expert exploration demonstrations~\cite{imitation_learning}. The expert demonstration is a sequence of image and action pairs taken on a route that efficiently covers a new environment. This could come from either an \textit{oracle policy} having full access to virtual environments or simply a \textit{human expert} in the real world.

\begin{figure*}[t]
    \centering
    \includegraphics[width=0.80\linewidth]{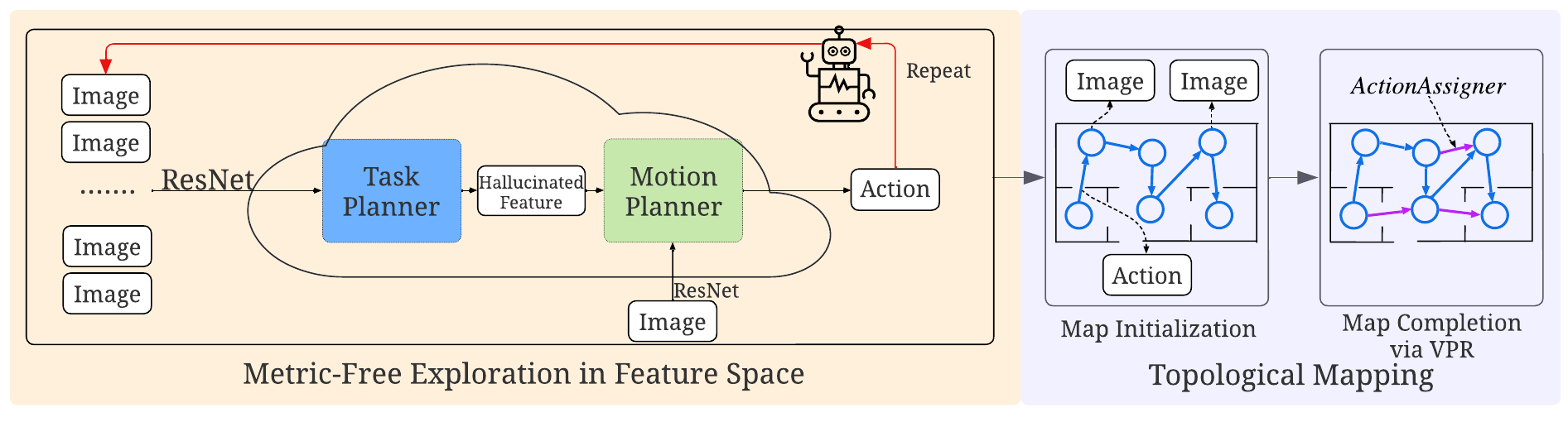}
    \vspace{-3mm}
    \caption{\textbf{Workflow of \textit{\acronym{}} and the following topological mapping.} The agent explores a novel environment by task and motion planning in feature space~(left); The following topological mapping completes the initial topological map by adding new edges via VPR and \textit{ActionAssigner}. }
    \label{fig:general_overview}
    \vspace{-5mm}
\end{figure*}

\textit{\acronym{}} follows the task and motion planning formalism~(TAMP) and entirely works in image feature space.
Its task planner, a two-layer LSTM~\cite{LSTM} network, conceives the next best goal feature to be explored by hallucination from the latest sequence of observed image features. Its motion planner, a simple multi-layer perceptron~(MLP), fuses the current and the hallucinated features to predict the best action moving the agent toward the hallucinated feature. Both the two planners are trained jointly by deep supervision~\cite{deeply_supervised_nets} of per-step feature hallucination and action prediction.
The trained \textit{\acronym{}} are deployed by iteratively calling the task and the motion planners to predict the next action.

\textit{\acronym{}} is designed for active topological mapping of unknown environments. During each exploration step, the topological map is updated by adding the latest image observation as a new node and the action on the new edge. We further adopt VLAD-based~\cite{VLAD} visual place recognition (VPR)~\cite{VPR_survey} for loop closing, adding additional new edges between image pairs that are temporally disjoint but spatially close. In the meantime, we train an \emph{ActionAssigner} to assign each VPR-added new edge with corresponding actions that move the agent from one node to the other. We call the above process as \textit{Topological Mapping}~(see Fig.~\ref{fig:general_overview}). Finally, the completed topological map efficiently represents environment connectivity and traversability. We can apply it to various robot tasks like visual navigation~\cite{savinov2018semiparametric}.

We demonstrate the advantage of \textit{\acronym{}} on both visual \textbf{exploration} and \textbf{navigation} tasks. We train it on Gibson~\cite{gibson_env} simulation dataset and test its \textbf{exploration} efficiency on both Gibson validation and MP3D~\cite{Matterport3D} dataset~(for zero-shot sim2sim generalization test). We further show its strong zero-shot sim2real generalization capability by directly deploying the Gibson-trained \textit{\acronym{}} to explore a real-world environment. For the navigation task, we run experiments on both Gibson~\cite{gibson_env} and MP3D~\cite{Matterport3D} dataset with the topological map built by \textit{\acronym{}}. 

In summary, our contributions are listed as follows:

\begin{itemize}
\item We propose \textit{\acronym{}} for efficient metric-free visual exploration based on task and motion planning entirely in an image feature space.

\item We train \textit{\acronym{}} via deeply-supervised imitation through joint feature hallucination and action prediction, whose importance is shown in our ablation study.

\item Through experiments on both exploration and navigation tasks, we show the efficiency and strong sim2sim/sim2real generalization capability of \textit{\acronym{}}.
\end{itemize}

%% file: RelatedWork.tex
\section{Related Work}

\textbf{Topological Map in Exploration and Navigation.} Inspired by the animal and human psychology~\cite{tolman1948cognitive}, a large amount of work has recently proposed to build topological map to represent an environment~\cite{graphtopoexplore,Murphy08ICRA,neural_topomap,beeching2020learning,savinov2018semiparametric,VisualGraphMem_ICCV21,MRTopoMap,Savarese-RSS-19}. They use the topological map for tasks such as navigation~\cite{neural_topomap,learn2explore_iclr20,savinov2018semiparametric,VisualGraphMem_ICCV21,Savarese-RSS-19}, exploration~\cite{learn2explore_iclr20,graphtopoexplore,Murphy08ICRA,savinov2018semiparametric,MRTopoMap,TSGM} and planning~\cite{beeching2020learning}. To build the topological map, they combine various sensors such as RGB image, depth map~\cite{TSGM,Savarese-RSS-19}, pose~\cite{neural_topomap,learn2explore_iclr20,beeching2020learning} and even LiDAR scanner~\cite{MRTopoMap,graphtopoexplore}. Some of them further adopt data-hungry and computation-demanding Reinforcement Learning~(RL) techniques to train the model to construct the topological map~\cite{neural_topomap,learn2explore_iclr20,VisualGraphMem_ICCV21}. Kwon \textit{et al.}~\cite{VisualGraphMem_ICCV21} combine imitation learning~(IL) and RL to train the model. Some of these methods~\cite{neural_topomap,learn2explore_iclr20,beeching2020learning} involve metric information to construct the topological map. N.~Savinov~\textit{et. al.}~\cite{savinov2018semiparametric} use the random walk to construct the topological map, which inevitably leads to an inefficient topological map. TSGM~\cite{TSGM} jointly adds surrounding objects during topological map construction. Unlike these prior works, our \textit{\acronym{}} is completely metric-free and simple in experimental configuration~(just RGB image, much smaller expert demonstration size).

\textbf{Hallucinating Future Feature.} The idea of hallucinating future latent features has been discussed in other application domains. Previous work has utilized this idea of visual anticipation in video prediction/human action prediction~\cite{16Vondrick,17Zeng,20Chang,21Fernando,Suris2021LearningTP}, and researchers have applied similar ideas to robot motion and path planning~\cite{Jain2016RecurrentNN, Koppula2016, Carlone2019, Park2016}. As stated in~\cite{16Vondrick,17Zeng,Suris2021LearningTP}, visual features in the latent space provide an efficient way to encode semantic/high-level information of scenes, allowing us to do planning in the latent space, which is considered more computationally efficient when dealing with high-dimensional data as input~\cite{Lippi2020,Ichter2019}. Different from previous robotics work, we take advantage of this efficient representation by adding deep supervision when anticipating the next visual feature, which was computationally intractable if we were to operate at the pixel level.

\textbf{Deeply-Supervised Learning} has been extensively explored~\cite{deeply_supervised_nets,knowledge_synergy,li2017deep,li2018deep} during the past several years. The main idea is to add extra supervision to various intermediate layers of a deep neural network in order to more effectively train deeper neural networks. In our work, we adopt a similar idea to deeply supervise the training of feature hallucination and action generation.

\begin{figure*}[t]
    \centering
    \includegraphics[width=0.8\linewidth]{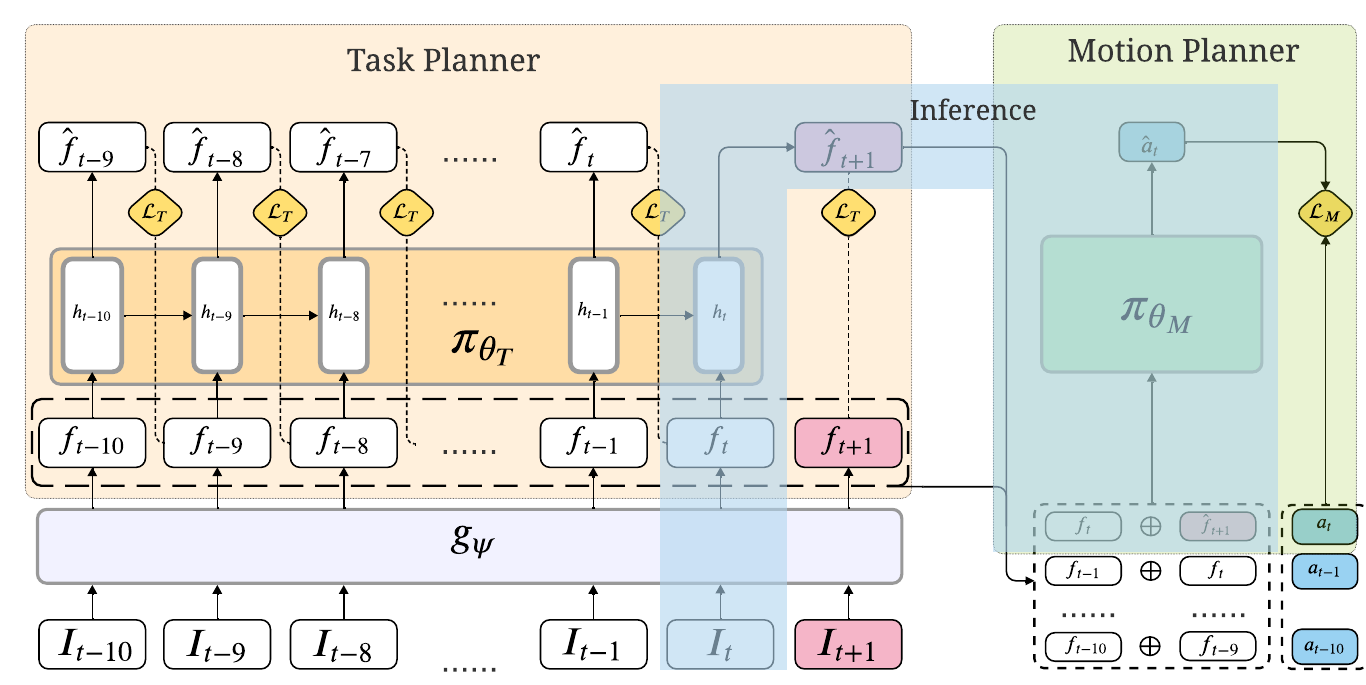}
    \caption{\textbf{Training and inference for task and motion imitation.} Feature extractor $g_\psi$ takes image $I_t$ as input and generates the corresponding feature vector $f_t$. \textit{TaskPlanner} $\pi_{\theta_T}$ is a recurrent neural network (RNN) consuming a sequence of features $\{ f_{t-10}, \cdots, f_t\}$ to hallucinate the next best feature to visit $\hat{f}_{t+1}$. \textit{MotionPlanner} $\pi_{\theta_M}$ consumes the concatenation (denoted by $\bigoplus$) of ${f}_{t}$ and $\hat{f}_{t+1}$ and generates the action to move the agent towards the hallucinated feature. During training, we supervise all the intermediate outputs including the intermediate hallucinated features $\{ \hat{f}_{t-9}, \cdots, \hat{f}_{t} \}$ and the intermediate actions $\{ \hat{a}_{t-10}, \cdots, \hat{a}_{t-1} \}$, in addition to the final output $\hat{f}_{t+1}$ and $\hat{a}_t$. During inference, current observation $I_t$ is firstly encoded and fed into $\pi_{\theta_T}$ to hallucinate $\hat{f}_{t+1}$, and then $\hat{f}_{t+1}$ combined with the ${f}_{t}$ is fed into $\pi_{\theta_M}$ for motion planning. $\mathcal{L}_T$ is $L_2$ loss and $\mathcal{L}_M$ is cross entropy loss (the subscripts $T$ and $M$ denote \textbf{T}ask and \textbf{M}otion respectively). $h_t$ denotes the hidden state of RNN.} 
    \label{fig:pipeline}
    \vspace{-5mm}
\end{figure*}

\textbf{Task and Motion Planning.} Task and motion planning (TAMP) divides a robotic planning problem into high-level task allocation (task planning) and low-level action for task execution (motion planning). This hierarchical framework is adopted in many robotic tasks such as manipulation \cite{chitnis2016guided,mcdonald2022guided} exploration~\cite{Cao-RSS-21} and navigation \cite{lo2018petlon,thomas2021mptp}. Such a framework allows us to leverage high-level information about the scenes to tackle challenges in local control techniques~\cite{bansal2019-lb-wayptnav}. In this work, to perform active topological mapping of a novel environment, the agent firstly reasons at the highest level about the regions to navigate: hallucinate the next best feature point to visit. Afterward, the agent takes an action to get to the target feature. The whole procedure is totally implemented in feature space without any metric information.

\textbf{Imitation Learning} aims to mimic human behavior or expert demonstrations for a given specific task~\cite{imitation_learning,il_legged,il_planning}. The agent is trained to perform tasks by directly observing demonstrations~\cite{il_legged,il_planning}. In our work, the expert demonstration is a set of image-action pair sequences that an agent would observe along a route that efficiently covers an environment. It is widely accessible in either real-world or simulated environments~(e.g. from human experts or maps of environments). 

%% file: mainsec1_activeexp.tex
\section{Topological Exploration in Feature Space}
\label{active_explore}

Our topological map is represented by a graph $\mathcal{G} = (\mathcal{I}, \mathcal{A})$, where the graph nodes denoted by $\mathcal{I}$ is a set of RGB panoramic image observations collected by the agent at different locations $\mathcal{I} = \{I_1, I_2, \cdots, I_N\}$ (where $N$ denotes the number of nodes), and the edges denoted by $\mathcal{A}$  is composed of a set of actions $a_{(I_i, I_j)} \in \mathcal{A}$ which moves an agent between the two spatially adjacent observations $I_i$ and $I_j$. Each RGB panoramic image is of size $256\times 512$, and the action space consists of three basic actions: \texttt{move\_forward}, \texttt{turn\_left}, and \texttt{turn\_right}. Our visual exploration aims at maximizing the topological map coverage over an environment given a certain step budget $N$. The coverage of the topological map denoted by $\mathcal{C}$ is defined as the total area in the map that is known to be traversable or non-traversable. Mathematically, let $\pi_\theta$ denote the policy network parameterized by $\theta$, $a_t$ denote the action taken at step $t$, and $\Delta \mathcal{C} (a_t)$ denote the gain in coverage introduced by taking action $a_t$, the following objective function is optimized to obtain the optimal exploration policy $\pi^*$:
\vspace{-2mm}
\begin{equation}\label{eq:general_objective}
    \pi^* = \arg\max_{\pi_{\theta}} \underset{a_t \sim \pi_{\theta}}{\mathbb{E}}(\sum_{t=1}^{N}\Delta \mathcal{C}(a_t)) \ .
\end{equation}

\textbf{Learning from expert demonstrations.} In literature, most works solve Eq.~(\ref{eq:general_objective}) by reinforcement learning to maximize the reward~\cite{chen2018learning,active_slam}, such solutions are not only data-hungry but also require complicated training involving metric information. Differently, we adopt imitation learning~\cite{imitation_learning} to let our policy network $\pi_\theta$ mimic the output of the expert policy $\tilde{\pi}$ which could come from either an \textit{oracle policy} having full access to virtual environments or simply a \textit{human expert} in real world~(more discussion is in Sec.~\ref{expconfig_sec}). Hence, our objective is to minimize the difference between our policy network and the expert policy:
\begin{equation}\label{eq:our_objective}
    \pi^* = \arg\min_{\pi_{\theta}} \mathcal{L}(\pi_\theta,\tilde{\pi}) \ ,
\end{equation}
where $\mathcal{L}$ measures the discrepancy between two policies. We propose the task and motion imitation in feature space to solve Eq.~(\ref{eq:our_objective}) which will be introduced in the following (see Fig.\,\ref{fig:pipeline}). We respectively introduce the feature extraction (\ref{subsec:feature_extraction}), the policy network $\pi_\theta$ composed of a \textit{TaskPlanner} denoted by $\pi_{\theta_T}$ (\ref{subsec:task_planner}) as well as a \textit{MotionPlanner} denoted by $\pi_{\theta_M}$ (\ref{subsec:motion_planner}), and the deeply-supervised learning strategy (\ref{subsec:learning}).

\subsection{Image Feature Extraction}
\label{subsec:feature_extraction}
We firstly encode each visual observation $I_t \in \mathcal{I} (t=1,2,...,N)$ with a feature extractor $g_\psi$ parameterized by $\psi$ which uses the ImageNet\,\cite{ILSVRC15} pre-trained ResNet18 backbone\,\cite{resnet18}. The feature embedding $f_t \in \mathbb{R}^{d} (d=512)$ is obtained by $f_t = g_\psi (I_t)$, (see Fig.\,\ref{fig:pipeline}). Note that $g_\psi$ is jointly optimized with the task planner $\pi_{\theta_T}$ as well as the \textit{MotionPlanner} $\pi_{\theta_M}$ via imitation learning.

\subsection{Task Planner for Next Best Feature Hallucination}\label{subsec:task_planner}
\textit{TaskPlanner} $\pi_{\theta_T}$ parameterized by $\theta_T$  takes the most recent $m$-step visual features $\mathcal{F}=\{f_{t-m}, \cdots, f_{t}\}$ as input, and learns to hallucinate the next best feature to visit which is denoted by $\hat{f}_{t+1}$, see Fig.~\ref{fig:pipeline}. In specific, $\pi_{\theta_T}$ is a two-layer LSTM\,\cite{LSTM}: 
\begin{equation}\label{eq:task_planner}
    \hat{f}_{t+1} = \pi_{\theta_T}(f_{t-m}, \cdots, f_{t} | \theta_T) \ .
\end{equation}

To save computation, $\pi_{\theta_T}$ only takes the most recent $m$-step features as input and we empirically find that $m=10$ achieves good performance. In other words, \textit{TaskPlanner} is only equipped with a short-term scene memory, and it tries to extend the feature space as quickly as possible in order to guide the agent to perform efficient exploration.  Essentially, \textit{TaskPlanner} is planning in the feature space. This efficient representation of the environment enables us to deploy deep supervision strategy introduced in Section \ref{subsec:learning}.

\subsection{Motion Planner for Action Generation}\label{subsec:motion_planner}
\textit{MotionPlanner} $\pi_{\theta_M}$ parameterized by $\theta_M$ takes the hallucinated feature $\hat{f}_{t+1}$ and the current feature $f_{t}$ as input, and outputs the action taking the agent towards the hallucinated goal~(see Fig.~\ref{fig:pipeline}). Specifically, $\pi_{\theta_M}$ is a multi-layer-perceptron (MLP) taking the concatenation of two features as input to classify the action:
\begin{equation}\label{eq:motion_planner}
    \hat{a}_{t} = \pi_{\theta_M}(\hat{f}_{t+1}, f_{t} | \theta_M) \ .
\end{equation}

\subsection{Deeply-Supervised Imitation Learning Strategy}\label{subsec:learning}

Our imitation pipeline is shown in Fig.\,\ref{fig:pipeline}. Given an expert exploration demonstration including a sequence of images and the corresponding expert actions $\mathcal{{E}}=\{\{{I}_1, {a}_1\}, \{{I}_2, {a}_2\}, \cdots, \{{I}_N, {a}_N\}\}$, we adopt the deeply-supervised learning strategy~\cite{deeply_supervised_nets} to jointly optimize the feature extractor $g_\psi$, task planner $\pi_{\theta_T}$, and \textit{MotionPlanner} $\pi_{\theta_M}$. Ultimately, our objective in Eq.~(\ref{eq:our_objective}) becomes,
\begin{equation}\label{eq:our_objective1}
    \min_{\psi,\theta_T,\theta_M} \sum_{t=1}^{N-1}\mathcal{L}_T(\hat{f}_{t+1},{f}_{t+1}) + \sum_{t=1}^{N}\mathcal{L}_M(\hat{a}_t, {a}_t) \ ,
\end{equation}
where $\mathcal{L}_T$ is $L_2$ loss to measure the discrepancy between two features, and $\mathcal{L}_M$ is cross-entropy loss to make the model imitate the expert action. The desired target feature ${f}_{t+1}$ is obtained by ${f}_{t+1} = g_\psi({I}_{t+1})$ (${I}_{t+1}$ is obtained from the expert demonstration  $\mathcal{E}$), the desired action ${a}_{t}$ is also read from $\mathcal{E}$, the hallucinated feature $\hat{f}_{t+1}$ is calculated by Eq.~(\ref{eq:task_planner}), and the generated action $\hat{a}_t$ is computed by Eq.~(\ref{eq:motion_planner}). For each training iteration, we randomly clip  $m+1$ observations and the corresponding $m$ actions from an expert exploration\,($m=10$ and $N\gg m$), and feed them to $g_\psi$, $\pi_{\theta_T}$, and $\pi_{\theta_M}$. During exploration, we iteratively take the latest $m$ image observations as input, after which we first call task planner $\pi_{\theta_T}$ to hallucinate the next best feature and then motion planner $\pi_{\theta_M}$ to predict the next best action taking the agent to the hallucinated feature accordingly. By constantly executing the predicted action, the agent efficiently explores an environment. 

The whole pipeline is shown in Fig.~\ref{fig:pipeline}, in which we deeply supervise all intermediate output. Specifically, in \textit{TaskPlanner}, instead of simply hallucinating the next best feature, we simultaneously hallucinate all intermediate feature for each step and supervise all hallucinations by truly image observations. In \textit{MotionPlanner}, we deeply supervise the action prediction in a similar fashion. We show by experiment that such deeply-supervised learning strategy~\cite{deeply_supervised_nets} endows the agent with more powerful exploration capability.

%% file: mainsec2_topomapping.tex
\subsection{VPR for Loop Closing}

The topological map $\mathcal{G} = (\mathcal{I}, \mathcal{A})$ initialized by the active exploration experience in Sec.~\ref{active_explore} is unidirectionally connected in the temporal axis. Each node~(a panoramic RGB image observation) is just connected with its preceding node and next node, failing to reflect the nodes' spatial adjacency. We propose to further complete the initial map $\mathcal{G}$ by adding edges to any two unconnected nodes if they possess a high visual similarity. In this work, we adopt VLAD-based visual place recognition~(VPR)~\cite{VLAD, netVLAD} to measure the ``visual similarity'' between two nodes.

Specifically, given $N$ unidirectionally connected image nodes collected during exploration in a room scene, we extract the local SIFT~\cite{SIFT} feature for each image. Then we get the global VLAD~\cite{VLAD} descriptor for each image by first clustering all SIFT features with K-Means~\cite{K-means} into $k$ centroids~(in our case $k=16$), and then stacking the residuals between the local SIFT features and centroids. After VLAD descriptors construction, we store all VLAD features into a ball tree~\cite{liuBallTree,bansal2019-lb-wayptnav} with leaf size 60. Then we can query each image's top-N ``most visually similar'' images from the corresponding ball tree, the node pairs whose similarity score is below a threshold~(in our case 1.15) are added edges. 

\begin{table}[t]
    \centering
        \caption{Total Loop Closing Time for Topological Mapping on Gibson 14 Room Scenes. Hardware: 10 cores of Intel Xeon Platinum 8268, 32GB RAM, and an SSD. SPTM requires an Nvidia A100 GPU.}
    \begin{tabular}{c|c}
    \toprule
      Methods   & Total Time Spent \\
      \hline
      SPTM~\cite{savinov2018semiparametric}  & $\approx 2.1$ hrs \\
      \textbf{VLAD-Based VPR (Ours)} & $\approx 0.2$ hrs\\
      \bottomrule
    \end{tabular}
    \label{tab:vpr_time}
\end{table}

It is worth noting that our VLAD-based VPR is more efficient for loop closing than SPTM~\cite{savinov2018semiparametric} which uses a binary classification network that requires exhaustive pairwise checking to detect loops. We report the average loop closing time of the two methods on all the 14 Gibson test rooms in Table~\ref{tab:vpr_time}, showing VLAD-based VPR's speed advantage.

Apart from the VPR, we train a model named \textit{ActionAssigner} to assign an action list to each new edge. The architecture of \textit{ActionAssigner} is similar to \textit{MotionPlanner}, except that \textit{ActionAssigner} predicts a sequence of actions with two node features as input, while \textit{MotionPlanner} is a one-step action predictor~(predict just one action).

After topological mapping, the completed topological map represents a room scene through the edges between nodes and the actions corresponding to each edge. It reflects both spatial adjacency and traversability of the room scene so that it can be used for navigation tasks. Given the image observations for the start and goal positions, we localize them on a topological map via the same VPR procedure. Once localized, we apply Dijkstra's algorithm~\cite{dijkstra} to find the shortest path between the two nodes. We can then navigate the agent from the start position to the goal without metric information.

%% file: experiment.tex
\section{Experiments}

\begin{figure*}[t]
    \centering
    \includegraphics[width=0.9\linewidth]{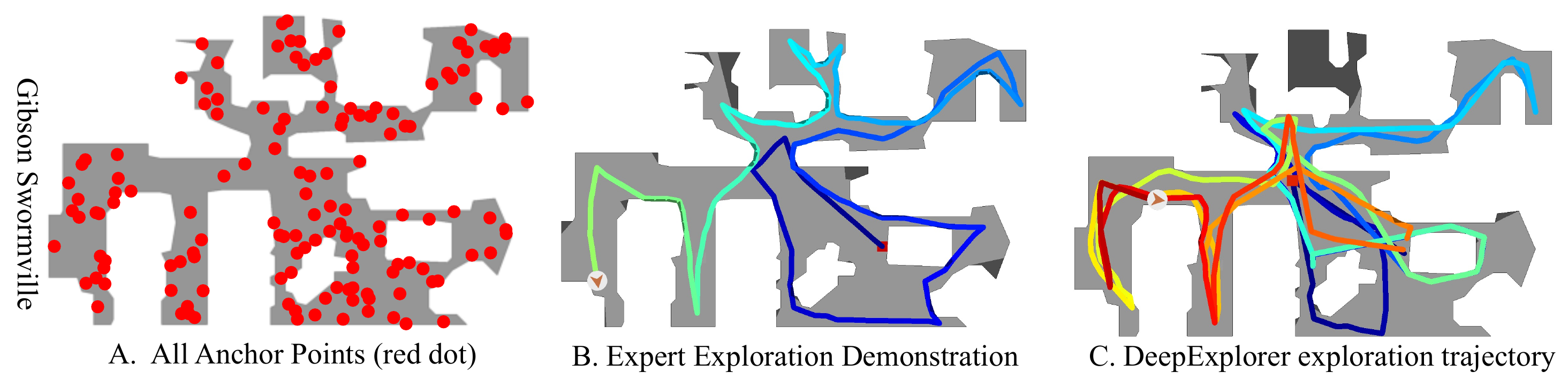}
    \vspace{-3mm}
    \caption{\textbf{Expert Demonstration and \textit{\acronym{}} exploration visualization}. \textbf{A}: all potential anchor points in the Gibson Swormville scene. \textbf{B}: expert selectively traverses a subset of the anchor points~(by merging spatial close anchor points) by iteratively reaching the next unvisited closest anchor points so as to create expert demonstrations. \textbf{C}: \textit{\acronym{}} exploration trajectory with a 1000-step budget, with the model learned from expert demonstrations.}
    \label{fig:expert_explore}
    \vspace{-3mm}
\end{figure*}

We test \textit{\acronym{}} on two datasets: Gibson~\cite{gibson_env} and Matterport3D~(MP3D)~\cite{Matterport3D} dataset on Habitat-lab platform~\cite{habitat19iccv}. The two datasets are collected in real indoor spaces by 3D scanning and reconstruction methods. The agents can be equipped with multi-modality sensors to perform various robotic tasks. The average room size of MP3D~(100 $m^2$) is much larger than that of Gibson~($[14 m^2, 85m^2]$).

\begin{table*}[t]
\scriptsize
  \centering
    \caption{Coverage Ratio over 1000-step budget. Top three performances are highlighted by \textbf{\textcolor{red}{red}}, \textbf{\textcolor{Green}{green}}, and \textbf{\textcolor{blue}{blue}} color, respectively.}
  \begin{tabular}{c|c|c|c|cc|cc}
  \hline
    \multirow{2}{*}{Method Description}&\multirow{2}{*}{Method} & \multirow{2}{*}{Sensor Used} & \multirow{2}{*}{\#Train Imgs}  & \multicolumn{2}{c|}{Gibson Val} & \multicolumn{2}{c}{\shortstack{Domain Generalization \\ MP3D Test}}  \\
    \cline{5-8}
   & & & & \%Cov. & Cov. ($m^2$)~& \%Cov. & Cov. ($m^2$) \\
\hline
Non-learning Based & RandomWalk~(used by SPTM~\cite{savinov2018semiparametric}) & No & No & 0.501 & 22.268 & 0.301 & 40.121   \\
\hline
\multirow{5}{*}{RL w/ Metric Input/Estimates} & RL + 3LConv~\cite{eval_metric} & \multirow{6}{*}{RGB, Depth, Pose} & 10~M & 0.737 & 22.838 & 0.332 & 47.758 \\
&RL+ResNet18 &  & 10~M & 0.747 & 23.188 & 0.341 & 49.175\\
&RL+ResNet18+AuxDepth~\cite{learning2complex} &  & 10~M & 0.779 & 24.467 & 0.356 & 51.959\\
&RL+ResNet18+ProjDepth~\cite{chen2018learning} &  & 10~M & 0.789 & 24.863 & 0.378 & 54.775\\
&OccAnt~\cite{ramakrishnan2020occant} &  & 1.5-2~M & \textbf{\textcolor{Green}{0.935}} & \textbf{\textcolor{blue}{31.712}} & 0.500 & 71.121 \\
&ANS~\cite{learn2explore_iclr20} &  & 10~M & \textbf{\textcolor{red}{0.948}} & \textbf{\textcolor{Green}{32.701}} & 0.521 & 73.281 \\
\hline
\multirow{4}{*}{\acronym{} Model Variants} & \acronym\_NoDeepSup & \multirow{6}{*}{RGB only} & \multirow{6}{*}{0.45~M} & 0.768 & 26.671 & 0.292 & 37.163 \\
 & \acronym\_NoFeatDeepSup &  &  & 0.912 & 35.151 & 0.620 & 104.499 \\
 & \acronym\_NoActDeepSup  &  &  & 0.900 & 33.922 & 0.600 & 102.122 \\
 & \acronym\_LSTMActRegu &  &  & 0.914 & 35.238 & 0.610 & 101.734 \\
 & \acronym\_withHistory &  &  & 0.917 & 35.331 & \textbf{\textcolor{blue}{0.618}} & \textbf{\textcolor{blue}{102.302}} \\
 & \acronym\_NoFeatHallu &  &  & 0.907 & 34.563 & 0.589 & 99.091 \\
 \cline{1-2}\cline{5-8}
\multirow{2}{*}{Deeply Supervised Imitation}&\textbf{\acronym} &  &  &  0.918 &35.274 & \textbf{\textcolor{Green}{0.642}} & \textbf{\textcolor{Green}{109.057}} \\
&\textbf{\acronym~(0.30m/$30^\circ$)} & &  & \textbf{\textcolor{blue}{0.927}} & \textbf{\textcolor{red}{37.731}} & \textbf{\textcolor{red}{0.656}} & \textbf{\textcolor{red}{117.993}} \\
\hline
  \end{tabular}
  \label{tab:coverage_ratio}
\end{table*}

We run experiments on two tasks: (1) \textbf{autonomous exploration} proposed by Chen \textit{et al.}~\cite{chen2018learning}, in which the target is to maximize an environment coverage within a fixed step budget~(1000-step budget following~\cite{learn2explore_iclr20}), and (2) \textbf{image-goal navigation} where the agent uses the constructed topological map to navigate from current observation to target observation. Regarding the exploration, we employ two evaluation metrics: (1) coverage ratio which is the percentage of the covered area over all navigable area, and (2) absolute covered area~($m^2$). We exactly follow the setting by ANS~\cite{learn2explore_iclr20} that a point is covered by the agent if it lies within the agent's field-of-view and is less than $3.2m$ away. Regarding the navigation, we adopt two evaluation metrics: shortest path length~(SPL) and success rate~(Succ. Rate)~\cite{eval_metric}. We again follow ANS~\cite{learn2explore_iclr20} to train \acronym{} on Gibson training dataset~(72 scenes), and test \acronym{} on Gibson validation dataset~(14 scenes) and MP3D test dataset~(18 scenes). Testing on the MP3D dataset helps to show \acronym's generalizability.

\subsection{Experiment Configuration}
\label{expconfig_sec}
\textbf{Exploration setup.} In exploration, we independently explore each scene 71 times, each time assigning the agent a random starting point~(created by a random seed number). We keep track of all the random seed numbers for result reproduction. We use the Habitat-lab sliding function so that the agent does not stop when it collides with a wall but instead slides along it. In order to generate the initial 10 steps required by \textit{\acronym}, we constantly let the agent execute \texttt{move\_forward} action. Once it collides with the wall, it randomly chooses \texttt{turn\_left} or \texttt{turn\_right} action to continue to explore. Afterward, we iteratively call \textit{TaskPlanner} and \textit{MotionPlanner} to efficiently explore the environment. During \textit{\acronym}-guided exploration, we allow the agent to actively detect its distance with surrounding obstacles or walls~(by using a distance sensor). When the agent's forward-looking distance to the closest obstacle or wall is less than 2-step distances and the \textit{\acronym{}} predicted action is \texttt{move\_forward}, we randomly choose either \texttt{turn\_left} and \texttt{turn\_right} to execute so as to avoid colliding with an obstacle. It is worth noting that using an obstacle avoidance scheme does not lead to unfair comparison because the comparing metric-based methods internally preserve a global metric map, which serves a similar purpose to help the agent avoid obstacles.

We experiment with two locomotion setups: the first one is with step-size 0.25~m and turning angle $10^{\circ}$, which follows the same setting established in~\cite{learn2explore_iclr20} for comparing with baseline methods in the exploration task. The second one is with step-size 0.30~m and turn-angle $30^{\circ}$. This setting helps us test \textit{\acronym}'s generalization capability under different locomotion configurations.

\textbf{Navigation setup.} In navigation, we encourage the agent to visit enough positions for each room scene. Specifically, the agent has collected 2,000 images per room on Gibson and 5,000 images per room on MP3D (2,000/5,000-step \textit{\acronym}-guided exploration).

\textbf{Expert demonstration generation.} For each room scene, we first sample multiple anchor points across the whole navigable area for each room scene. Then the agent starts at a random anchor point and iteratively walks to the next unvisited closest anchor point with minimal steps~(by calling Habitat \texttt{PathFollower} API) until all anchor points are traversed. At each step, we record the agent's action and panoramic RGB image. Please refer to Fig.~\ref{fig:expert_explore} for a visualization of this process. It is worth noting that our expert demonstration does not necessarily guarantee globally optimal exploration. The way we obtain the expert demonstration can be easily automated and scaled.

\textbf{Training details.} The network architectures for both \textit{TaskPlanner} and \textit{MotionPlanner} are given in Table~\ref{atm_network},\ref{actionassigner_network} in Appendix . In our implementation, the local observation sequence length is 10 (m=10) because we empirically found it to achieve a good performance-memory trade-off. We experimentally tested $m=20$ and got inferior performance. \textit{\acronym{}} network architecture is illustrated in Appendix~(parameter size is just 16~M). We train \textit{\acronym{}} with PyTorch~\cite{pytorch_package}. The optimizer is Adam~\cite{adam_optimizer} with an initial learning rate of 0.0005, but decays every 40 epochs with a decaying rate of 0.5. In total, we train 70 epochs. We train all the \textit{\acronym{}} variants with the same hyperparameter setting for a fair comparison.

\begin{figure*}[t]
    \centering
    \includegraphics[width=0.90\linewidth]{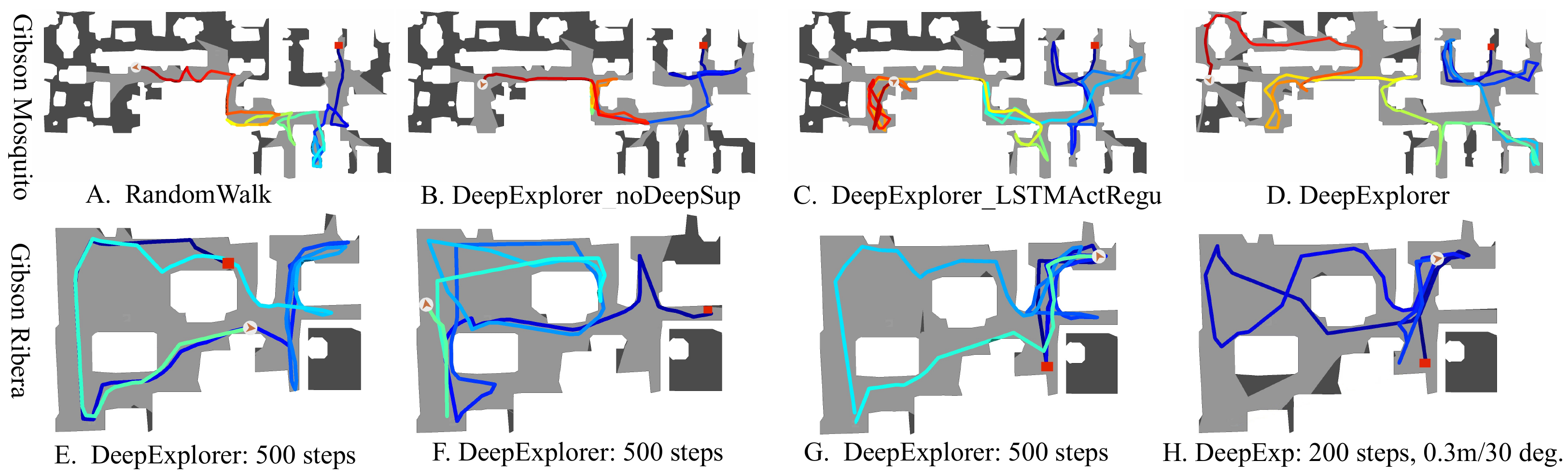}
    \vspace{-3mm}
    \caption{\textbf{Exploration trajectories Visualization}. Top row: various \textit{\acronym{}} variants exploration result~(1000-step budget) on Gibson \texttt{Mosquito} scene. Bottom row: exploration with different start positions~(E, F, G, 500-step budget, with agent step-size 0.25~m and turn-angle $10^\circ$). An agent with larger step size and turn angle~(0.3~m/$30^\circ$) achieves a similar coverage ratio with much smaller steps~(200 steps, F). The trajectory color evolving from cold\,(blue) to warm~(yellow) indicates the exploration chronological order.}
    \vspace{-4mm}
    \label{fig:traj_vis}
\end{figure*}

\begin{figure*}[h]
    \centering
    \includegraphics[width=0.95\linewidth]{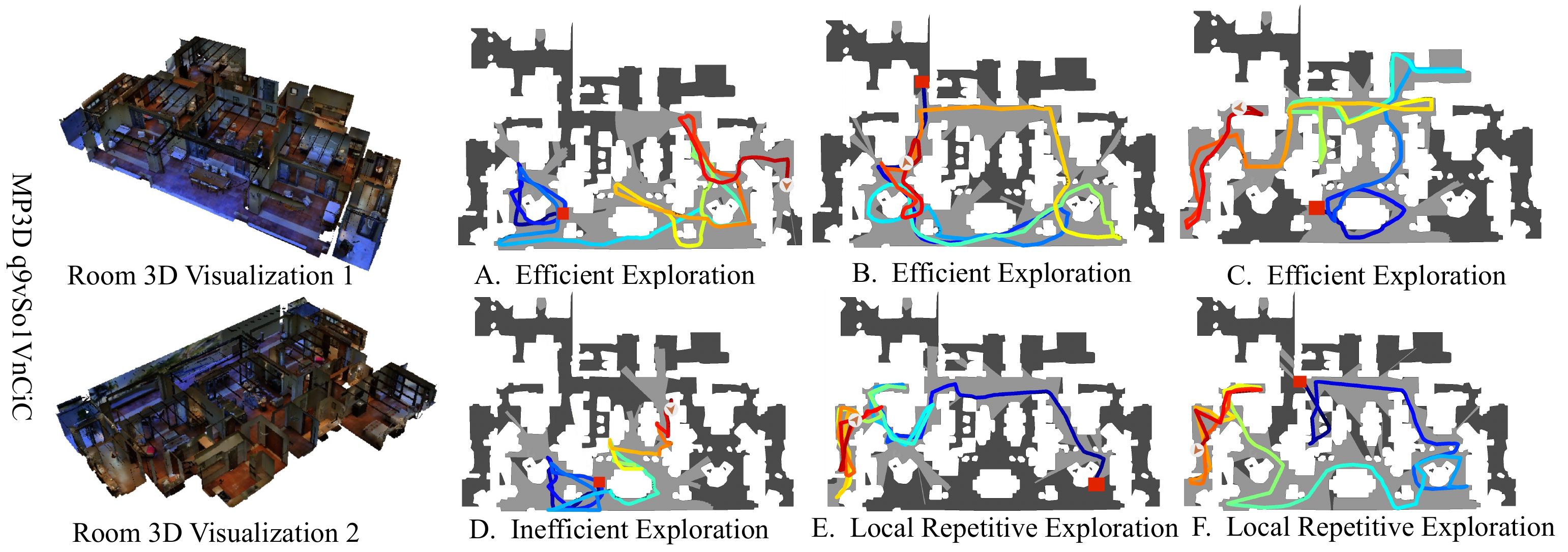}
    \caption{DeepExplorer exploration result on MP3D~\cite{Matterport3D} room scene \texttt{q9vSo1VnCiC}. We show both efficient exploration results ~(top row, sub-figure A, B, C). We also show relatively less-efficient exploration in sub-figure E and F, which are mainly due to local repetitive exploration. We further show an inefficient exploration example in sub-figure D. Two-room scene 3D visualization is given in the left-most subfigures. The agent exploration starting position is marked by a red rectangle patch.}
    \label{fig:mp3d_explore_rst}
\end{figure*}
\subsection{Comparison Methods}

For exploration task, we compare \textit{\acronym{}} with six RL-based methods: 1. \textbf{RL + 3LConv}: An RL Policy with 3 layer convolutional network~\cite{habitat19iccv}; 2. \textbf{RL + Res18}: RL Policy initialized with ResNet18~\cite{resnet18} and followed by GRU~\cite{GRU}; 3. \textbf{RL + Res18 + AuxDepth}: adapted from~\cite{learning2complex} which uses depth map prediction as an auxiliary task. The network architecture is the same as ANS~\cite{active_slam} with one extra deconvolution layer for depth prediction; 4. \textbf{RL + Res18 + ProjDepth} adapted from Chen \textit{et al.}~\cite{chen2018learning} who project the depth
image in an egocentric top-down in addition to the RGB image as input to the RL policy. 5. \textbf{ANS}~(Active Neural SLAM~\cite{learn2explore_iclr20}) jointly learns a local and global policy network to guide the agent to explore; 6. \textbf{OccAnt}~\cite{ramakrishnan2020occant}: takes RGB, depth, and camera as inputs to learn a 2D top-down occupancy map to help exploration. For ablation studies, we have following \textit{\acronym{}} variants:

\begin{enumerate}
    \item \textbf{RandomWalk} The agent randomly chooses an action to execute at each step. It serves as a baseline and helps us to know agent exploration capability without any active learning process. Please note that RandomWalk is also the SPTM~\cite{savinov2018semiparametric} exploration strategy.
    \item \textbf{\acronym\_NoDeepSup.} \textit{\acronym{}} without deeply-supervised learning. We remove LSTM per-step feature supervision in \textit{TaskPlanner} and neighboring frame action supervision in \textit{MotionPlanner}. In other words, we just keep the feature prediction and action classification between the latest step and the future step. It helps to test the necessity of involving a deeply-supervised learning strategy.
    \item \textbf{\acronym\_NoFeatDeepSup.} \textit{\acronym{}} without deeply-supervised learning in the feature space. We remove LSTM per-step feature supervision in \textit{TaskPlanner} but keep the neighboring frame action supervision in \textit{MotionPlanner}. This means no $\mathcal{L}_T$ but only $\mathcal{L}_M$. Together with \textbf{\acronym\_NoDeepSup} and \textbf{\acronym\_NoActDeepSup}, it helps to test the necessity of deploying deep supervision in both task and motion planning.
    \item \textbf{\acronym\_NoActDeepSup.} \textit{\acronym{}} without deeply-supervised learning regarding action prediction. We remove the neighboring frame action supervision in \textit{MotionPlanner} but keep the LSTM per-frame feature supervision in \textit{TaskPlanner}. In other words, there is no $\mathcal{L}_M$ but only $\mathcal{L}_T$. Together with \textbf{\acronym\_NoDeepSup} and \textbf{\acronym\_NoFeatDeepSup} It helps to test the necessity of deploying deep supervision in both task and motion planning.
    \item \textbf{\acronym\_LSTMActRegu.} \textit{TaskPlanner} hallucinates the next-best feature at each step to deeply supervise the whole framework in the feature space. As an alternative, we can instead predict action instead at each step in \textit{TaskPlanner}. This \acronym{} variant helps us to figure out whether supervising each step of \textit{TaskPlanner} LSTM in feature space is helpful.
    \item \textbf{\acronym\_withHistory.} \textit{\acronym{}} is trained with only a short-memory\,(the latest $m$ steps observations). To validate the influence of long-term memory, we train a new \textit{\acronym{}} variant by adding extra historical information: we evenly extract 10 observations among all historically explored observations excluding the latest $m$ steps. After feeding them to ResNet18\,\cite{resnet18} to get their embedding, we simply use average pooling to get one 512-dimensional vector and feed it to \textit{TaskPlanner} LSTM as the hidden state input.
    \item \textbf{\acronym\_noFeatHallu.} We use the architecture of \textit{TaskPlanner} to directly predict the next action. It discards task planning in feature space but instead plans directly in action space. Its performance helps us to understand if the hallucinated feature is truly necessary.
    \item \textbf{\acronym~(0.30m/$30^\circ$).} This variant adopts a different locomotion protocol than the one used in ANS \cite{learn2explore_iclr20} and all other variants to demonstrate \textit{\acronym{}}'s robustness under different locomotion setups.

\end{enumerate}
Some visualizations of different \textit{\acronym} variants' exploration results can be found in Fig.~\ref{fig:traj_vis}.

\subsection{Evaluation Results on Exploration}
The quantitative results of the exploration task are shown in Table~\ref{tab:coverage_ratio}. We can observe from this table that \textit{\acronym{}} achieves comparable performance on the Gibson dataset with the best RL-based methods and best-performing result on the MP3D dataset by outperforming all RL-based methods significantly~(about $13\%$ coverage ratio and $40 m^2$ area improvement). Since the comparing RL-based methods~\cite{eval_metric,learning2complex,chen2018learning,learn2explore_iclr20} build the map in metric space and requires millions of training images, \acronym{} is desirable because (1) it provides a metric-free option for exploration, and (2) it is lightweight~(in terms of parameter size 16~M) and requires much less training data~(just about 0.45 million images, in contrast with 10 million images required by most RL-based methods). The room scenes in MP3D dataset are much more complex and larger than those in the Gibson dataset. They contain various physical impediments~(e.g., complex layout, furniture), and some rooms contain outdoor scenarios. Hence, \textbf{\acronym{} exhibits stronger zero-shot sim2sim generalizability in exploring novel scenes than RL-based methods}. Moreover, the performance gain is more obvious on both Gibson and MP3D datasets when we change the agent to a different locomotion setup~(from 0.25/$10^\circ$ to 0.30/$30^\circ$), which also shows \acronym{} is robust to different locomotion setups.

\begin{figure*}[t]
    \centering
    \includegraphics[width=0.80\linewidth]{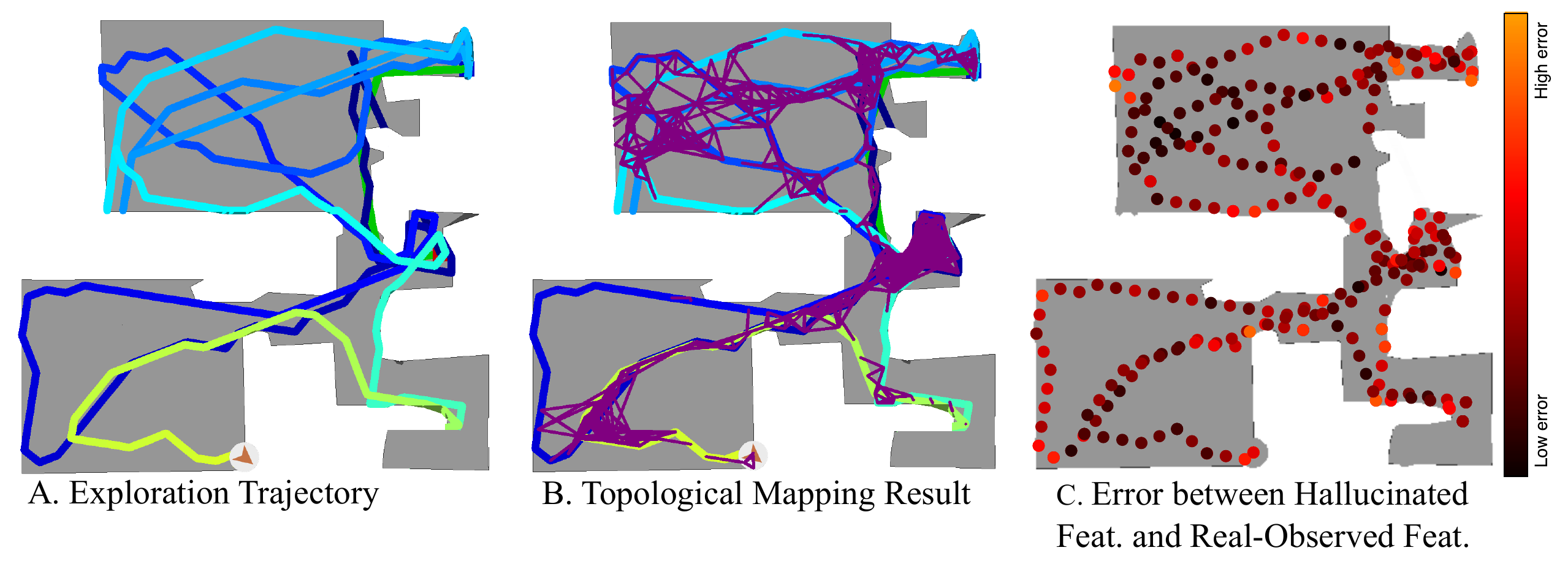}
    \vspace{-4mm}
    \caption{\textbf{Feature Visualization} \textbf{A}. The exploration trajectory~(blue to yellow, step size 0.25~m and turn-angle $30^\circ$) with a 500-step budget overlaid on top of the floor plan map. \textbf{B}. The spatially-adjacent panoramic images are connected~(purple lines) via VPR. \textbf{C}. The difference~(Euclidean distance in 512-d feature space) between real-observed features and hallucinated features. The darker the color, the lower the difference.}
    \label{fig:feat_vis}
\end{figure*}

On the Gibson dataset, \textit{\acronym{}} achieves a slightly lower coverage ratio than ANS~\cite{learn2explore_iclr20} but a higher average covered area. We find such performance difference is mainly caused by \textit{\acronym{}} stronger capability in exploring large areas than RL-based methods. In most cases, \textit{\acronym{}} actively reaches new areas within limited steps.

\textbf{Comparison with random exploration.} \textit{RandomWalk} serves as the baseline for our framework. It is also adopted by SPTM~\cite{savinov2018semiparametric} to build a topological map. It involves no learning procedure, and the agent randomly takes action at each step to explore an environment. From Table\,\ref{tab:coverage_ratio}, we can see that \textit{RandomWalk} dramatically reduces the exploration performance in terms of both coverage ratio and average coverage area. The inferior performance of \textit{RandomWalk} verifies the necessity of learning active exploration strategy in order to help the agent efficiently explore an environment. Figure~\ref{fig:traj_vis} demonstrates the qualitative comparison between \textit{RandomWalk} and \textit{\acronym{}} exploration result.

\textbf{Feature regularization and with history memory}. If we replace feature regularization involved in \textit{TaskPlanner} with action regularization~(\textit{\acronym\_LSTMActRegu}), we have observed more performance drop on MP3D than on Gibson dataset~($3\%$ versus $0.2\%$), which shows adopting feature regularization improves the generalizability  compared with action regularization. Moreover, introducing full history memory~(\textit{\acronym\_FullHistory}) to \textit{TaskPlanner}~(used as LSTM hidden state input) produces very similar results on the Gibson dataset, but significantly reduces the performance on MP3D dataset~(more than $2\%$ drop). It thus shows using historical memory tends to encourage \textit{\acronym{}} to overfit training data so that its generalizability is inevitably reduced.  We argue that such generalizability drop might lie in our over-simplified history memory modeling because we just evenly sample 10 nodes~(image observations) from all historically visited nodes, which might be too simple to represent the whole history memory, or even confuses \textit{TaskPlanner} if the agent has already explored many steps. A more elegant long-term history memory model remains to be explored.

\textbf{Deeply-supervised learning and joint task and motion imitation.} Removing deeply-supervised learning~(\textit{\acronym\_noDeepSup}, \textit{\acronym\_noFeatDeepSup}, \textit{\acronym\_noActDeepSup}) leads to performance drop on both Gibson and MP3D dataset, especially when both deep supervisions in \textit{TaskPlanner} and \textit{MotionPlanner} are both dropped. In the MP3D dataset, it can even lead to worse performance than \textit{RandomWalk}. It thus shows the necessity of deep supervision in both feature space~(\textit{TaskPlanner}) and action space~(\textit{MotionPlanner}). Meanwhile, \textit{\acronym\_noFeatHallu} leads to a significant performance drop on both Gibson and MP3D datasets. It thus attests to the advantage of our feature-space task and motion imitation strategy which jointly optimize \textit{TaskPlanner} for high-level task allocation and \textit{MotionPlanner} for low-level motion control.

We also visualize the comparison between \textit{\acronym{}} hallucinated next-step future feature and truly observed feature in Fig.~\ref{fig:feat_vis}~(C). We see that the hallucinated feature is more similar to the observed real feature when the agent is walking through a spacious area~(in other words, the agent mostly takes \texttt{move\_forward} action) than when the agent is walking along a room corner, against the wall or through a narrow pathway. This may be due to the learned \textit{TaskPlanner} most likely hallucinates feature moving the agent forward if the temporary egocentric environment allows. This also matches expert exploration experience because experts mostly prefer moving forward so as to explore as many areas as possible.

\begin{table*}[t]
  \centering
    \caption{Navigation Result. Top three performances are highlighted in \textbf{\textcolor{red}{red}}, \textbf{\textcolor{Green}{green}}, and \textbf{\textcolor{blue}{blue}} color, respectively. We collect 2000~(Gibson)/5000(MP3D) images with agent setup~(0.25m/$10^\circ$). `N/A' means `not available.'}
    \scriptsize
    \label{table:nav_rst}
    \vspace{-1mm}
  \begin{tabular}{c|cc|cc}
  \hline
    \multirow{2}{*}{Method}  & \multicolumn{2}{c|}{Gibson Val} & \multicolumn{2}{c}{\shortstack{Domain Generalization on MP3D Testset}} \\
    \cline{2-5}
    & Succ. Rate~($\uparrow$) & SPL~($\uparrow$) & Succ. Rate~($\uparrow$) & SPL~($\uparrow$)\\
\hline
RandomWalk & 0.027 & 0.021 & 0.010 & 0.010 \\
RL + Blind & 0.625 & 0.421 & 0.136 & 0.087 \\
RL + 3LConv + GRU\,\cite{eval_metric} & 0.550 & 0.406 & 0.102 & 0.080 \\
RL + ResNet18 + GRU & 0.561 & 0.422 & 0.160 & 0.125 \\
RL + ResNet18 + GRU + AuxDepth\,\cite{learning2complex} & 0.640 & 0.461 & 0.189 & 0.143 \\
RL + ResNet18 + GRU + ProjDepth\,\cite{chen2018learning} & 0.614 & 0.436 & 0.134 & 0.111 \\
IL + ResNet18 + GRU & 0.823 & 0.725 & \textbf{\textcolor{blue}{0.365}} & \textbf{\textcolor{blue}{0.318}} \\
SPTM~\cite{savinov2018semiparametric} & 0.510 & 0.381 & 0.240 & 0.203 \\
CMP~\cite{CMP} & 0.827 & \textbf{\textcolor{blue}{0.730}} & 0.320 & 0.270 \\
OccAnt~(RGB)~\cite{ramakrishnan2020occant} & \textbf{\textcolor{blue}{0.882}} & 0.712 & N/A & N/A \\
ANS~\cite{learn2explore_iclr20} & \textbf{\textcolor{Green}{0.951}} & \textbf{\textcolor{Green}{0.848}} & \textbf{\textcolor{Green}{0.593}} & \textbf{\textcolor{Green}{0.496}} \\
\hline
\acronym{}& \textbf{\textcolor{red}{0.957}} & \textbf{\textcolor{red}{0.859}} & \textbf{\textcolor{red}{0.733}} & \textbf{\textcolor{red}{0.619}} \\
\hline
  \end{tabular}
  \label{tab:navigation}
\end{table*}

\subsection{Evaluation Results on Navigation}

For the visual navigation task, we compare \textit{\acronym{}} with most of the methods compared in the exploration task. CMP~\cite{CMP} builds up a top-down belief map for joint planning and mapping. For OccAnt~\cite{ramakrishnan2020occant}, we just report its result with the model trained with RGB image~(so as to be directly comparable with \textit{\acronym}). For SPTM~\cite{savinov2018semiparametric}, we train all its navigation-relevant models on data obtained by \textit{\acronym}. The navigation result is given in Table~\ref{table:nav_rst}. We can see that \textit{\acronym{}} outperforms all comparing methods on the two datasets, with the largest performance gain on the MP3D dataset~(about $14\%$ Succ. Rate, $12\%$ SPL improvement). Hence, we can see that our \textit{\acronym}-built topological map can be used for image-goal-based visual navigation. More importantly, \textit{\acronym{}} shows satisfactory zero-shot sim2sim generalizability in navigation as well. In Fig.~\ref{fig:feat_vis}~(B), we can see VPR and \textit{ActionAssigner} successfully add new edges~(purple lines) for loop closing. The resulting topological map, after topological mapping, fully reflects environment connectivity and traversability.

\subsection{Zero-Shot Sim2Real Real-World Exploration}

 \textit{\acronym{}} is deployed and verified on a customized real-world robot. We set an Insta360 Pro 2 camera\footnote{\url{https://www.insta360.com/cn/product/insta360-pro2}} on an iRobot Create 2 robot\footnote{\url{https://edu.irobot.com/what-we-offer/create-robot}}~(the camera height is around $1.5~m$). Nvidia Jetson TX2\footnote{\url{https://www.nvidia.com/en-gb/autonomous-machines/embedded-systems/jetson-tx2/}} platform is used to launch the \textit{\acronym{}} model and control the robot. We directly deploy the model~(with step size $0.25m$ and turn-angle $10^{\circ}$) trained on the Gibson simulation dataset without any fine-tuning on the real-world dataset. The robot's physical configuration is as close to that of the simulation as possible. We adopt a LiDAR scanner for obstacle avoidance. The experiment environment is a large indoor multi-functional office building.

We find that \textit{\acronym{}} demonstrates strong zero-shot sim2real exploration results: the robot is capable of identifying obstacles and actively reaching the open navigable areas. The robot can traverse the entire hallway and enter the only open door (marked with a star in Fig.~\ref{fig:teasingfig}) and manage to exit it through the door after exploring it. The exploration trajectory is shown in Fig.~\ref{fig:teasingfig} and Fig.~\ref{fig:realworld_traj_redstar}~(in Appendix). The corresponding video can be found on the Github repository.

\subsection{Limitations}
In our experiment on the simulation datasets, we find that \textit{\acronym{}} sometimes leads to inefficient exploration in complex room environments as is shown in Fig.~\ref{fig:mp3d_explore_rst}~(bottom row), especially when the room layout is sophisticated and the navigable area is narrow. We hypothesize that this is partly due to the lack of full history memory of \textit{\acronym{}} that can steer the agent away from already-covered areas. Although we have tried one simple history memory mechanism~(\acronym\_withHistory), it still remains a future research topic to design a better history memory framework. 

In the zero-shot sim2real exploration experiment, we find that the agent often mistakes large glass walls for open doorways. We speculate that the lack of relevant data in the Gibson training dataset, which is entirely made of the household environment, leads to this failure. Extra measurement should be considered to handle such cases.

%% file: conclusion.tex
\section{Conclusion}
Our proposed \textit{\acronym} is capable of efficiently building a topological map by metric-free exploration to represent an environment. It entirely works in an image feature space to explore a new environment by jointly hallucinating the next step feature and predicting the appropriate action that best moves the agent to the feature. It is simple and lightweight as it just requires RGB images and the model size is small. It is trained via deeply-supervised imitation learning where the expert demonstration is easy to acquire and scale up. We show its strong zero-shot sim2sim and sim2real generalization capability by experiments in both large-scale and photo-realistic simulation environments and real-world environments. Future works include designing more elaborate historic memory modules and involving multi-modality sensors to further improve the performance of visual exploration and navigation.

%% file: suppmaterial.tex
\section*{Appendix}

\subsection{DeepExplorer Neural Network Architecture}
\label{nn_architecture}
\textit{DeepExplorer} neural network architecture is given in Table~\ref{atm_network}, it consists of ResNet18~(for image observation embedding), LSTM layer~\cite{LSTM}~(for \textit{TaskPlanner}) and  multi-layer perceptron~(MLP)~(for \textit{MotionPlanner}). Please note that \textit{DeepExplorer} is lightweight, its parameter size is just 16~M~(where M is million).

\begin{table}[h]
    \centering
    \small
    \caption{DeepExplorer network architecture illustration. The network consists of basic 2D image convolution layers, such as ResNet18, LSTM, and FC. The network is lightweight, the parameter size is just 16~M.}
    \scriptsize
    \begin{tabular}{|c|c|c|}
        \hline
        Layer Name & Filter Num  & Output Size \\ 
        \hline
        \multicolumn{3}{|c|}{\textbf{Image Embedding Layer}}\\
        \hline
        \multicolumn{3}{|c|}{Input: [10, 3, 256, 512]}\\
        \multicolumn{3}{|c|}{Embedding Network: ResNet18} \\
        \multicolumn{3}{|c|}{Embedding Size: [10, 512]}\\
        \hline
        \multicolumn{3}{|c|}{\textbf{Task Planner Network}}\\
        \hline
        LSTM & layers = 2, hidden size = 512 & [10, 512]\\
        % \hline
        Feat Prediction FC & in feat = 512, out feat = 512 & [10, 512] \\
        \hline
        \multicolumn{3}{|c|}{\textbf{Motion Planner Network}} \\
        \hline
        \multicolumn{3}{|c|}{Input: Feat [10, 1024], Action: [10]} \\
        \hline
        Feat Merge FC & in feat = 1024, out feat = 512 & [10, 512]\\
        Action Classification FC & in feat = 512, out feat = 3 & [10, 3] \\
        \hline
    \end{tabular}
    \label{atm_network}
\end{table}

\begin{table}[h]
    \centering
    \small
    \caption{\textit{ActionAssigner} Neural Network Architecture. The whole parameter size is $13.5$~M.}
    \scriptsize
    \begin{tabular}{|c|c|c|}
        \hline
        Layer Name & Filter Num  & Output Size \\ 
        \hline
        \multicolumn{3}{|c|}{\textbf{Image Embedding Layer}}\\
        \hline
        \multicolumn{3}{|c|}{Input: [2, 3, 256, 512]}\\
        \multicolumn{3}{|c|}{Embedding Network: ResNet18} \\
        \hline
        \multicolumn{3}{|c|}{\textbf{Feat. Merge Layer}}\\
        \hline
        \multicolumn{3}{|c|}{Concat. Size: [1, 1024]}\\
        \hline
        FC & in feat = 1024, out feat = 512 & [1, 512]\\
        \hline
        \multicolumn{3}{|c|}{\textbf{Action Predict Branch}}\\
        \hline
        head1 FC & in feat = 512, out feat = 128 & [1, 128] \\
        head2 FC & in feat = 512, out feat = 128 & [1, 128] \\
        head3 FC & in feat = 512, out feat = 128 & [1, 128] \\
        head4 FC & in feat = 512, out feat = 128 & [1, 128] \\
        head5 FC & in feat = 512, out feat = 128 & [1, 128] \\
        head6 FC & in feat = 512, out feat = 128 & [1, 128] \\
        \hline
        \multicolumn{3}{|c|}{\textbf{Action Predict}} \\
        \hline
        \multicolumn{3}{|c|}{Concat. Size: [1, 6, 128]} \\
        \hline
        BiLSTM & layers = 1, out feat = 128 & [1, 6, 128]\\
        Action Classify FC & in feat = 128, out feat = 3 & [1, 6, 3] \\
        \hline
    \end{tabular}
    \label{actionassigner_network}
\end{table}

\subsection{ActionAssigner Network Architecture}

The \textit{ActionAssigner} neural network is given in Table~\ref{actionassigner_network}. The \textit{ActionAssigner} also uses ResNet18~\cite{resnet18} as the image embedding module. Then it uses a sequence of multi-layer perceptron~(MLP) to predict multi-step actions separately~(step length is 6), each step independently predicts one action. So \textit{ActionAssigner} is a multi-label classification neural network. Bidirectional LSTM is applied to model mutual action dependency among different steps. The parameter size is $13.5$~M. We train \textit{ActionAssigner} with the same parameter setting as of \textit{TaskPlanner} and \textit{MotionPlanner}~(the network in Table~\ref{atm_network}). During training data preparation, if the action list length is smaller than 6, we pad \texttt{STOP} action to fill the length.

\subsection{Coverage Ratio Progression Comparison}

\begin{figure}[t]
    \centering
    \includegraphics[width=0.98\linewidth]{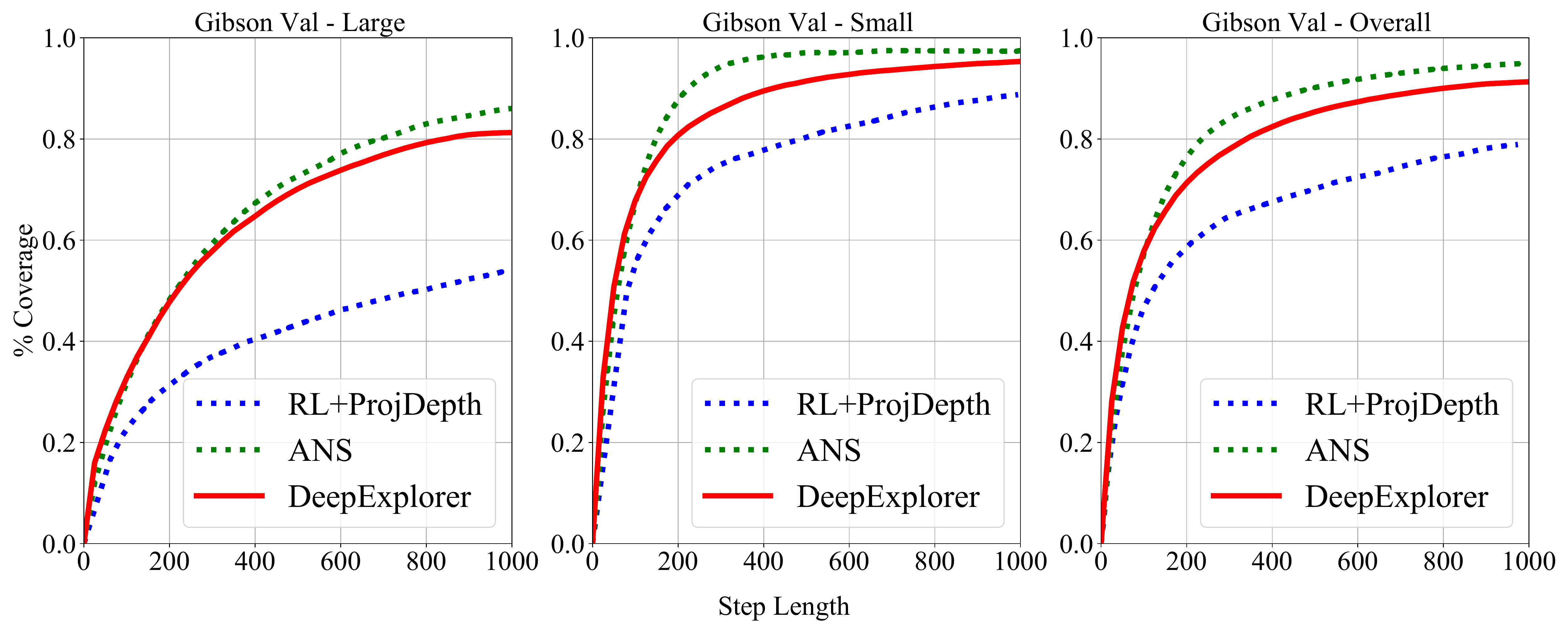}
    \caption{Coverage ratio variation curve comparison over 1000-step budget over large $> 50 m^2$, small $< 50 m^2$ and all~(average) room size, respectively.}
    \label{fig:coverage_ratio}
\end{figure}

We further provide the coverage ratio progression variation w.r.t. exploring steps comparison between \textit{DeepExplorer} and ANS~\cite{learn2explore_iclr20}, one RL-based method~(RL+ProjDepth) in Fig.~\ref{fig:coverage_ratio}. The comparison is based on Gibson validation dataset~\cite{gibson_env}, and we divide the room into \textit{Large}, \textit{Small}, and \textit{Overall} according to the room size. From this table, we can see that \textit{DeepExplorer} is capable of covering more area during the first 200 steps than ANS~\cite{learn2explore_iclr20} on large rooms~(which is verified by the more steep curve of \textit{DeepExplorer} over ANS~\cite{learn2explore_iclr20} and RL+ProjDepth, in the middle sub-figure).

\begin{figure}[t]
    \centering
    \includegraphics[width=0.90\linewidth]{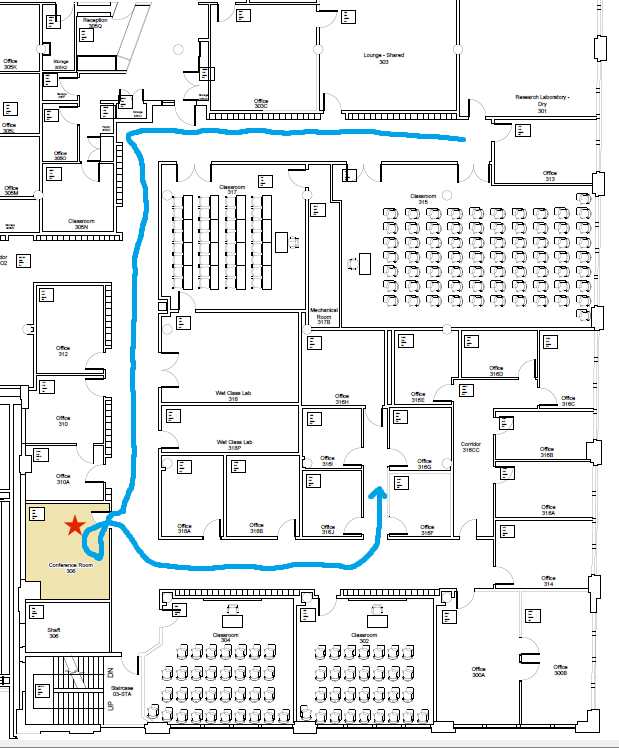}
    \caption{DeepExplorer exploration trajectory on zero-shot sim2real experiment. We can see that the agent can successfully find the navigable area along the corridor to efficiently explore more areas. It can also manage to enter and exit one conference room~(the area marked with a red star).}
    \label{fig:realworld_traj_redstar}
\end{figure}

\subsection{Zero-Shot Sim2Real Exploration Discussion}

We provide two exploration videos in the supplementary folder. One video demonstrates the successful exploration~(with the exploration trajectory shown in Fig.~\ref{fig:realworld_traj_redstar}), and the other video shows one unsuccessful exploration case in which the agent mixes the glass walls with an open area.

The agent we used for real-world exploration contains large actuation noise, so the actual angle it has turned may be different from our configuration~($10^\circ$). Sometimes its actual executed turn angle can be as large as $20\circ$ or even $30^\circ$, especially in the conference room where the floor is overlaid with carpet~(yellow color and red star marked area in Fig.~\ref{fig:realworld_traj_redstar}). The existence of actuation noise explains the non-smoothness between adjacent frames in our provided video, especially when the agent entered the conference room.

Although the actuation noise is caused by the agent, we find our trained \textit{DeepExplorer} model can predict the appropriate actions to mitigate the actuation noise impact. For example, when the agent has turned a larger angle than the configuration~(e.g. turn left), \textit{DeepExplorer} can predict a contrary action~(e.g. turn right) to the correct agent.